%% file: top.tex
\documentclass[letterpaper]{article} 
\usepackage{aaai2026}  
\usepackage{times}  
\usepackage{helvet}  
\usepackage{courier}  
\usepackage[hyphens]{url}  
\usepackage{graphicx} 
\usepackage{natbib}  
\usepackage{caption} 
\usepackage{amsmath}
\usepackage{amssymb}
\usepackage{xcolor}
\usepackage{enumitem}

\usepackage{algorithm}
\usepackage{algorithmic}

\usepackage{amsmath}
\usepackage{amssymb}
\usepackage{xcolor}
\usepackage{enumitem}
\usepackage{multirow} 
\usepackage{booktabs}
\usepackage{subcaption}
\usepackage{tabularx}
\usepackage{tikz}
\usepackage{pgfplots} 
\pgfplotsset{compat=1.18}
\usepackage{float}
\usepackage{makecell}
\usepackage{fancyhdr}
\usetikzlibrary{arrows.meta, positioning, decorations.pathreplacing, shapes, circuits.logic.US}

\newcommand{\ourmethod}{CASCAD}

\usepackage{newfloat}
\usepackage{listings}
\DeclareCaptionStyle{ruled}{labelfont=normalfont,labelsep=colon,strut=off} 
\floatstyle{ruled}
\newfloat{listing}{tb}{lst}{}
\floatname{listing}{Listing}
\nocopyright

\title{Circuit-Aware SAT Solving: Guiding CDCL via Conditional Probabilities}
\author{
    Jiaying Zhu$^{1, 2}$\equalcontrib,
    Ziyang Zheng$^{1, 2}$\equalcontrib,
    Zhengyuan Shi$^{1, 2}$\equalcontrib,
    Yalun Cai$^{1, 2}$,
    Qiang Xu$^{1, 2}$\thanks{Corresponding author: Qiang Xu}
}
\affiliations{
    $^1$Department of Computer Science and Engineering, The Chinese University of Hong Kong, Hong Kong SAR \\
    $^2$National Center of Technology Innovation for EDA, Nanjing, China \\
    \texttt{\{jyzhu24, zyzheng23, zyshi21, ylcai25, qxu\}@cse.cuhk.edu.hk}
}

\usepackage{bibentry}

\begin{document}

\maketitle

\begin{abstract}
Circuit Satisfiability (CSAT) plays a pivotal role in Electronic Design Automation. 
The standard workflow for solving CSAT problems converts circuits into Conjunctive Normal Form (CNF) and employs generic SAT solvers powered by Conflict-Driven Clause Learning (CDCL). However, this process inherently discards rich structural and functional information, leading to suboptimal solver performance. To address this limitation, we introduce \textbf{CASCAD}, a novel circuit-aware SAT solving framework that directly leverages circuit-level conditional probabilities computed via Graph Neural Networks (GNNs). By explicitly modeling gate-level conditional probabilities, CASCAD dynamically guides two critical CDCL heuristics—variable phase selection and clause management—to significantly enhance solver efficiency. Extensive evaluations on challenging real-world Logical Equivalence Checking(LEC) benchmarks 
demonstrate that CASCAD reduces solving times by up to 10× compared to state-of-the-art CNF-based approaches, achieving an additional 23.5\% runtime reduction via our probability-guided clause filtering strategy. Our results underscore the importance of preserving circuit-level structural insights within SAT solvers, providing a robust foundation for future improvements in SAT-solving efficiency and EDA tool design.
\end{abstract}


\input{01_Intro_Stone}

\input{02_Related}

\input{03_Method}
\input{04_Exp}
\input{05_CSAT}

\input{06_Con}

\clearpage
\newpage

\bibliography{ref}

\clearpage
\newpage
\input{07_Appendix}

\end{document}

%% file: 01_Intro_Stone.tex
\section{Introduction}


Circuit Satisfiability (CSAT), a prominent variant of Boolean Satisfiability (SAT), is fundamental to numerous critical tasks in Electronic Design Automation (EDA), including Logic Equivalence Checking (LEC)~\cite{Goldberg_Prasad_Brayton_2001} and Automatic Test Pattern Generation (ATPG)~\cite{Stephan_Brayton_Sangiovanni-Vincentelli_1996}. Traditionally, the prevailing approach to solving CSAT instances involves translating circuit representations into flat Conjunctive Normal Form (CNF) formulas and employing Conflict-Driven Clause Learning (CDCL)-based SAT solvers such as Kissat~\cite{kissat_2024} and Glucose~\cite{glucose_2018}. CDCL algorithms dynamically determine variable assignments and prune search spaces efficiently~\cite{marques2021conflict}. Despite their powerful heuristics and optimization techniques, CNF-based solvers inherently discard crucial structural and functional insights originally embedded in circuit representations, often leading to suboptimal performance in practical solving scenarios.

Recognizing this limitation, recent efforts have focused on leveraging native circuit structures to improve SAT-solving efficiency. However, most existing methods rely on static analysis or preprocessing techniques, such as logic simplifications and structural transformations~\cite{07sat, 25dac}, which lack dynamic guidance during solver execution. Although some learning-based approaches integrate graph embeddings and circuit features directly into SAT solving~\cite{amizadeh2018learning, shi2023deepgate2, shi2024deepgate3}, they predominantly utilize static graph features and struggle to adapt to evolving solver states. As a result, the dynamic use of circuit-level intelligence in SAT solving remains largely underexplored.

In this work, we identify the central obstacle to effective circuit-aware SAT solving as \textit{the disconnect between static circuit structures and the inherently dynamic nature of CDCL's reasoning process}. CDCL solvers operate through a sequence of state-dependent decisions, whereas static circuit analyses typically cannot reflect the solver’s evolving needs. To bridge this fundamental gap, we introduce a probabilistic reframing of the CSAT solving problem. Specifically, we posit that the conditional probabilities of unassigned circuit variables, given the solver’s prior decisions, directly capture the dynamic essence of CDCL reasoning.

\begin{figure*}
    \centering
    \includegraphics[width=0.95\linewidth]{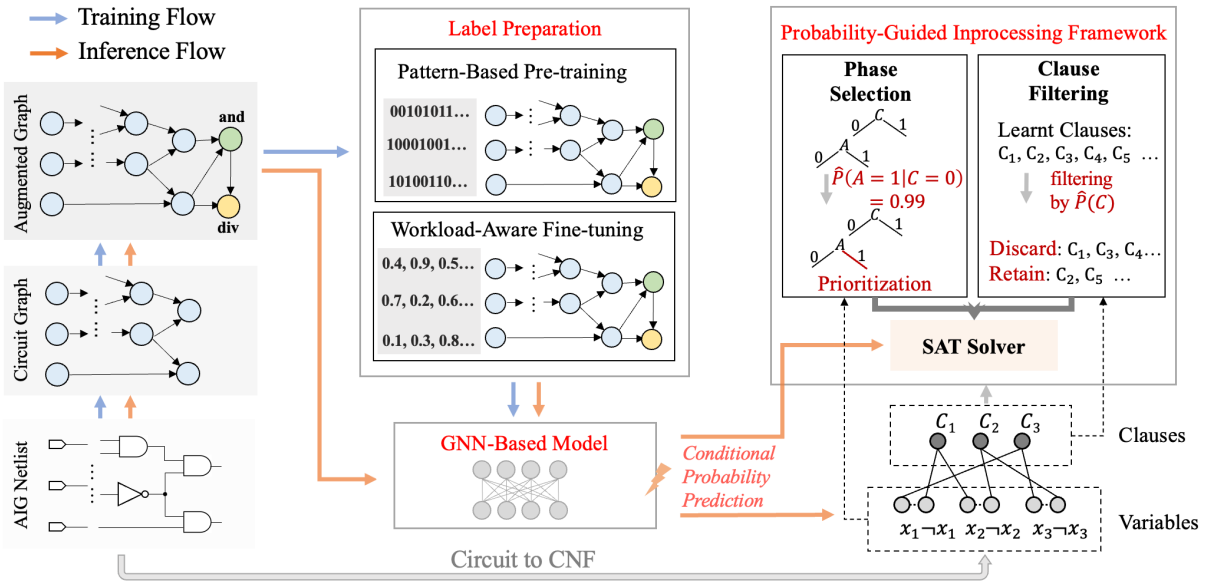}
    \caption{Framework of \ourmethod. We convert the original circuit into a DAG representation, augmented with virtual \texttt{and} and \texttt{div} gates. In the training pipeline, we employ pattern-based pre-training followed by workload-aware fine-tuning to train the model for conditional probability prediction. During inference, the predicted conditional probabilities are used to guide SAT solving, specifically in the inprocessing stages of phase selection and clause filtering. 
    }
    \label{fig:overview}
\end{figure*}

Based on this key insight, we propose \textbf{CASCAD} (\underline{C}ircuit-\underline{A}ware SAT \underline{S}olving via \underline{C}ondition\underline{A}l probability-guided \underline{D}ecisions), a novel SAT-solving framework that employs a Graph Neural Network (GNN) to dynamically estimate gate-level conditional probabilities directly from circuit structures. These probabilities are seamlessly integrated into two essential CDCL heuristics: \emph{variable phase selection} and \emph{clause management} (see Figure \ref{fig:overview}). By guiding these heuristics through real-time probability estimations, CASCAD maintains structural circuit intelligence throughout the solving process, significantly improving solver performance.

We extensively validate CASCAD on industry-standard EDA benchmarks.
Our results demonstrate that probability-guided phase selection alone reduces solving times by up to 90\% on challenging SAT instances derived from logic equivalence checking tasks. Furthermore, by employing a novel probability-based clause filtering strategy, CASCAD achieves an additional 23.5\% reduction in total solving time on benchmarks from logic equivalence checking scenarios.

In summary, our primary contributions include:
\begin{itemize}
    \item Introducing the first learning-based in-processing heuristics specifically designed for CSAT solving, effectively bridging static circuit representations and the dynamic CDCL solving process.
    \item Developing a highly accurate GNN-based probabilistic model that predicts gate-level conditional probabilities, achieving validation accuracy of 96.69\%.
    \item Demonstrating substantial empirical improvements by integrating the GNN-based conditional probability guidance into state-of-the-art SAT solvers, resulting in significant efficiency gains in real-world EDA applications.
\end{itemize}

Our approach not only advances the theoretical understanding of leveraging structural information in SAT solving but also provides a robust framework for future solver enhancements and EDA tool designs.

%% file: 02_Related.tex
\section{Related Work}

\subsection{Circuit Representation Learning}
Function-aware representation learning has become a critical subfield in Electronic Design Automation (EDA), reflecting the broader trend in AI toward learning unified embeddings that can serve multiple downstream tasks. This area can be categorized into two main approaches: predictive models and contrastive models.
In predictive models, DeepGate~\cite{shi2023deepgate2,shi2024deepgate3,zheng2025deepgate4} and DeepCell~\cite{shi2025deepcell} leverage asynchronous message passing nerual network with representation disentanglement techniques to generate separate embeddings for functionality and structure, with pretraining across various EDA benchmarks. PolarGate~\cite{PolarGate} further refines functional embeddings by incorporating ambipolar device principles.
In contrastive models, FGNN~\cite{wang2022functionality,wang2024fgnn2} employs contrastive learning to align circuit embeddings based on functional similarity. Additionally, MGVGA~\cite{wu2025circuit}, CircuitFusion~\cite{fang2025circuitfusion}, and NetTAG~\cite{fang2025nettag} apply contrastive learning across multi-modal circuits to obtain function-invariant embeddings.
In this paper, we focus on predicting exact conditional probabilities, for which we choose the state-of-the-art predictive model, DeepGate4, as backbone.
\subsection{Circuit Satisfiability}

The \textbf{Boolean Satisfiability Problem (SAT)} is a canonical NP-complete problem that determinies whether a given propositional formula is satisfiable.
As a variant of SAT, the \textbf{Circuit Satisfiability Problem (CSAT, Circuit SAT)} asks whether there exists an input assignment that makes the output of a given Boolean circuit evaluate to true. 
Several circuit-based solvers have been developed to address the Circuit SAT problem, aiming to operate directly on circuit format and leverage the structural information. For instance, NIMO~\cite{nimo} employs advanced circuit-level Boolean constraint propagation techniques to enhance conflict detection and decision-making. Similarly, QuteSAT~\cite{qutesat} incorporates conflict-driven learning strategies optimized for solving complex circuit topologies. Despite these innovations, such solvers generally offer only basic functionality and still fall short of the performance achieved by state-of-the-art CNF-based solvers on large-scale benchmarks.

Modern SAT solvers, such as Kissat~\cite{kissat_2024}, and CaDiCaL~\cite{fleury2020cadical}, have demonstrated remarkable success in handling CNF-based instances. 
The de-facto standard workflow for solving CSAT problems is to translate circuit into the CNF formulas to be processed by highly-optimized SAT solvers.
However, their performance on circuit-based problems is severely bottlenecked by the \textit{circuit-to-CNF transformation}, which disrupts the structural properties of circuits and produces solver-unfriendly representations. Previous efforts, such as applying EDA-driven circuit optimization techniques~\cite{07sat, 25dac} and extracting XOR logic on raw circuit ~\cite{cai2025xsat} to reformat the problem into solver-friendly representations before solving, have shown attractive improvements. However, these approaches remain static—they have no impact on solving heuristics and are inactive during the \textit{inprocessing phase}, a critical period where search decisions and analysis routines are dynamically adjusted. As a result, the rich structural intelligence of circuits remains untapped precisely when it could be most beneficial.


%% file: 03_Method.tex
\section{Methodology}

\subsection{Problem Definition}



\label{subsec:problem-definition}






During CDCL solving, the CNF formula evolves dynamically through decision making, conflict analysis, and clause management routines. Two core heuristics—\textit{phase selection} and \textit{clause filtering}—play a central role in guiding the search and maintaining solver efficiency. Phase selection determines polarity (true/false) of variable assignments during the search process, aiming to minimize conflicts and accelerate convergence. Meanwhile, clause filtering focuses on evaluating and managing learned clauses derived from conflict analysis, retaining high-quality ones that improve propagation efficiency while discarding low-value clauses to reduce memory overhead.

We reformulate both heuristics as conditional probability prediction tasks over circuit structures.

\subsubsection{Phase Selection}  

Given a Boolean circuit \(\mathcal{C}\) with primary output (PO), we formulate phase selection as a probability inference problem over the circuit graph.    
Let \( s \in V \) denote a variable corresponding to a gate in the circuit. At each decision step \( t \), the solver selects a variable \( s \in V \) and estimates the conditional probability:
$ P(s = 1 \mid \text{PO} = 1) $
which measures the likelihood that setting \( s = 1 \) helps satisfy the output. 
By leveraging these probabilities to guide phase assignments, the solver implicitly \textit{favors value choices that are more aligned with satisfying assignments}—often leading to stronger propagation and fewer conflicts.

A threshold \( \tau \in (0, 0.5) \) governs the phase assignment:

\begin{equation}
    v_s = 
    \begin{cases}
        0, & \text{if } P(s = 1 \mid \text{PO} = 1) < \tau, \\
        1, & \text{if } P(s = 1 \mid \text{PO} = 1) > 1 - \tau, \\
        \text{default}, & \text{otherwise}. 
    \end{cases}
\end{equation}

This probability inference strategy biases phase decisions toward values more likely to satisfy downstream logic, improving propagation and reducing conflicts.

\subsubsection{Clause Management}  
Existing metrics for clause quality like LBD and clause length rely solely on CNF-level features, ignoring structural insights from the original circuit, thus hindering performance on circuit-derived problems.

We propose a new metric, \textit{clause probability}, to evaluate the quality of learned clauses by estimating their likelihood of being satisfied under structural semantics. Consider a learned clause \( C = l_1 \lor l_2 \lor \dots \lor l_k \), which is semantically equivalent to a \(k\)-input OR gate over its literals. The probability that this clause is satisfied can be expressed as:

\begin{equation}
    P(C) = P(l_1 \lor l_2 \lor \dots \lor l_k).
\end{equation}

where \( P(l_i) \) is the estimated probability of literal \( l_i \) being true, derived from circuit-level signal estimation.

This probabilistic view provides an intuitive interpretation: a clause with high probability \( P(C) \) is likely to be satisfied easily, implying that it encodes \textit{a weak constraint and has limited impact on restricting the search space}. In contrast, a clause with low \( P(C) \) represents a strong constraint that is harder to satisfy, hence \textit{has higher potential to prune large infeasible regions in the search space}. Therefore, we regard clauses with lower \( P(C) \) as more informative and prioritize them during inprocessing.

Concretely, during periodic clause database elimination, the solver evaluates \( P(C) \) for all learned clauses and retains those with the lowest probabilities. This filtering mechanism allows the solver to preserve high-quality constraints that are more aligned with the underlying circuit semantics. 

We adopt the \textsc{DeepGate4} framework~\cite{zheng2025deepgate4} as our encoder \(E\), which generates structure and function embeddings for each gate in the circuit graph \(\mathcal{G}\). 
We use these embeddings to estimate conditional probabilities that guide inprocessing heuristics, enabling more efficient, probability-aware Circuit SAT solving.



\subsection{Graph Construction}
\label{subsec:dataset generation}

\begin{figure}
    \centering
    \includegraphics[width=\linewidth]{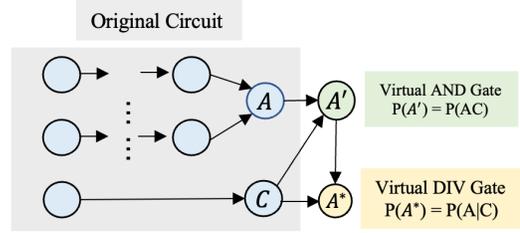}
    \caption{Augmented graph with virtual \texttt{and} and \texttt{div} gates.}
    \label{fig:dataset}
\end{figure}

To enable probability-aware learning on circuit graphs, we construct directed acyclic graphs (DAGs) where each node represents a logical gate, and further augment with virtual gates for better probability modeling, as shown in Figure \ref{fig:dataset}.

\subsubsection{DAG Construction with \texttt{and} and \texttt{not} Gates}
Following prior work~\cite{shi2023deepgate2,shi2024deepgate3,zheng2025deepgate4,PolarGate}, we begin by constructing a standard DAG representation of the input circuit. Each logic gate is represented as a node, and edges represent signal propagation between gates. Specifically, for a given circuit graph $\mathcal{G}$, we construct a DAG comprising two basic gate types: \texttt{and} gates for binary conjunctions and \texttt{not} gates for logic inversion.

We adopt the aggregator $Aggr_{and}$ and $Aggr_{not}$ for \texttt{and} gates and \texttt{not} gates respectively, following DeepGate4 framework\cite{zheng2025deepgate4}.

\subsubsection{Virtual \texttt{and} Gates for Joint Probabilities}
To expose joint probabilities directly in the graph, we insert \textbf{virtual \texttt{and} gates} in $\mathcal{G}$: for any node $A$ conditioned on $C$, we add
\begin{align}
    A_{joint} = A \land C, 
\end{align}
so the model can observe the joint probability $P(A \land C)$ directly, denoted as $P(A_{joint})$, without intermediate arithmetic computations. When encode the virtual \texttt{and} gate, we directly adopt the $Aggr_{and}$ to get the embedding and predict the joint probability as:
\begin{align}
    h^{A_{joint}}=Aggr(h^A,h^C), \ 
    \hat{P}(A_{joint}) = \varphi(h^{A_{joint}}),
\end{align}
where $\varphi$ is a 3-layer MLPs.
\subsubsection{Virtual \texttt{div} Gates for Conditional Probabilities}

As joint probabilities can be modeled via virtual \texttt{and} gates, computing conditional probabilities such as
\begin{align}
P(A \mid C) = \frac{P(A \land C)}{P(C)} = \frac{P(A_{joint})}{P(C)} 
\label{eq:conditional_prob}
\end{align}

can lead to \textit{amplified errors}, especially when $P(C)$ is small. Details are shown in Appendix \ref{app:div}.

Nodes with extremely low probabilities—referred to as \textbf{polar} nodes (i.e., those with truth-table probabilities below $0.1$)—are empirically harder to learn and tend to amplify prediction errors. These nodes require special attention to ensure accurate and reliable modeling.


To address this, we introduce \textbf{virtual \texttt{div} gates} to represent conditional probabilities directly within the graph. Each such gate takes two inputs: the numerator, which is a virtual \texttt{and} gate representing the joint event $A_{\text{joint}} = A \land C$, and the denominator, which corresponds to the condition node $C$.

Then with aggregator $Aggr_{div}$ designed for \texttt{div} gate, we first get the embedding with:
\begin{align}
h^{A_{cond}} =  Aggr_{div}(h^{A_{joint}},h^{C}).
\end{align}
Then predict the probability with task head $\varphi$:
\begin{align}
    \hat{P}(A_{cond})= \varphi(h^{A_{cond}})
\end{align}

This design enables the model to predict $P(A \mid C)$ directly from graph context, reducing error sensitivity to small denominators.

\subsection{Two-Stage Training Strategy}

We train the network with a two-stage training strategy:

    \subsubsection{1. Pattern-Based Pre-training.}
    
    \begin{table}[htbp]
    \centering
    \small
    {
      \begin{tabular}{l|cccc|ccc}
        \toprule
        Pattern ID & $PI_1$ & $PI_2$ & $PI_3$ & $PI_4$ & Node $x$ & Node $y$ & Node $z$ \\
        \midrule
        0 & 0 & 0 & 0 & 0 & 0 & 0 & 1 \\
        1 & 0 & 0 & 1 & 0 & 1 & 1 & 0 \\
        2 & 0 & 1 & 0 & 1 & 0 & 0 & 0 \\
        3 & 0 & 1 & 1 & 0 & 0 & 0 & 1 \\
        4 & 1 & 0 & 0 & 1 & 1 & 1 & 1 \\
        5 & 1 & 0 & 1 & 1 & 1 & 1 & 0 \\
        6 & 1 & 1 & 0 & 0 & 1 & 1 & 1 \\
        7 & 1 & 1 & 1 & 1 & 0 & 0 & 0 \\
        \bottomrule
      \end{tabular}
    }
    \caption{Example of simulation patterns and probability.}
    \label{tab:patterns}
    \end{table}


    Previous works~\cite{shi2023deepgate2,shi2024deepgate3,zheng2025deepgate4,PolarGate} initialize PI embeddings using input probabilities, however, it will lead to information distortion.
    As shown in Table~\ref{tab:patterns}, distinct input subsets (0–3 and 4–7) may share the same empirical PI probability (0.5), despite differing in actual assignments. This illustrates a key limitation: probabilities are coarse-grained and may fail to capture fine-grained circuit behavior.

    To address this, we use 100 random patterns per minibatch to compute fine-grained probability features, which are fed into the model. This supervision enables our model to capture both statistical trends and functional semantics.
    
    Suppose the circuit contains \(n\) nodes in total, where the first \(m\) nodes are designated as primary inputs (PIs).
    For each primary input (PI) node, a 100-bit simulation trace is encoded into an initial embedding using an autoencoder. Embeddings for all other nodes are then propagated level-by-level through the circuit. 

    The stage-1 training loss is defined as:
    \begin{equation}
    \begin{aligned}
    \mathcal{L}_{\text{stage1}} =\ 
    & w_1 \cdot \frac{1}{n - m} \sum_{i=m+1}^{n} L1(P_i - \hat{P}_i) \\
    & +\ w_2 \cdot \frac{1}{m} \sum_{j=1}^{m} L1(P_j - \hat{P}_j)
    \end{aligned}
    \end{equation}

    Here, the L1 Loss function is defined as:
    \begin{equation}
    \text{L1}(a, b) = |a - b|
    \end{equation}
    where \(P_i\) denotes the ground-truth probability of the \(i\)-th node, and \(\hat{P}_i\) is the predicted value. The second term focuses on PI nodes, encouraging accurate reconstruction of their probabilities, since their embeddings are initialized from actual simulation data. 

    \subsubsection{2. Workload-Aware Fine-tuning.}

    Although training on batches of 100 patterns helps capture fine-grained circuit behaviors, our ultimate goal is to generalize to the statistical properties of large-scale pattern distributions. 
    
    To bridge this gap and enhance data diversity, we perform 200 simulations, each using 100 random patterns with designated PI workloads drawn from $\{0.1, 0.2, \dots, 0.9\}$(detailed in Appendix \ref{app:workload}). For each PI, the 100-bit trace in a simulation is averaged into a single probability, resulting in a 200-dimensional vector that captures its behavior under diverse input distributions. This vector is then passed through an MLP to produce the PIs' initial embedding. 
    For internal nodes, ground-truth probabilities are computed by aggregating predictions over all \(200 \times 100 = 20{,}000\) input patterns, serving as training targets in stage-2.

    The overall loss for stage-2 is defined as:
    \begin{equation}
    \begin{aligned}
    \mathcal{L}_{\text{stage2}} =\ 
    & w_1 \cdot \frac{1}{|\mathcal{S}_\text{all}|} \sum_{i \in \mathcal{S}_\text{all}} L1(P_i - \hat{P}_i) \\
    & + w_2 \cdot \frac{1}{|\mathcal{S}_\text{div}|} \sum_{j \in \mathcal{S}_\text{div}} L1(P_j - \hat{P}_j) \\
    & +\ w_3 \cdot \frac{1}{|\mathcal{S}_\text{polar}|} \sum_{k \in \mathcal{S}_\text{polar}} L1(P_k - \hat{P}_k)
    \end{aligned}
    \end{equation}
    Here, $\mathcal{S}\text{all}$ denotes all internal nodes in the circuit, $\mathcal{S}\text{div}$ refers to \texttt{div} nodes, and $\mathcal{S}_\text{polar}$ includes polarized nodes with low probability (\(P < 0.1\)).

    Higher weights \(w_2\) and \(w_3\) prioritize structurally and semantically critical nodes: \texttt{div} nodes reflect key conditional probabilities, while polarized nodes are harder to predict but crucial for tasks like Circuit SAT.
    


\subsection{Multiple Condition Probability}

\begin{figure}
    \centering
    \includegraphics[width=\linewidth]{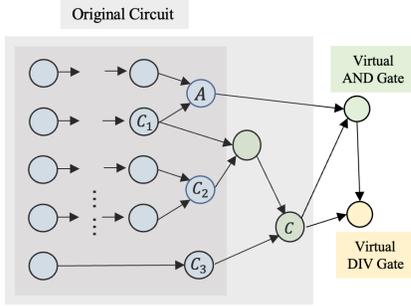}
    \caption{Example of conditional probability calculation for three conditions}
    \label{fig:5cond}
\end{figure}

Through Equation \ref{eq:conditional_prob}, our model can estimate the probability of a target node conditioned on a single other node. However, in practical Circuit SAT scenarios, it is often more useful to infer the most likely value of a node given that \textbf{multiple other nodes} have already been assigned.

As illustrated in Figure~\ref{fig:5cond}, when multiple condition nodes (denoted as $C_1, C_2, \ldots, C_k$) are involved, we combine them using a chain of two-input \texttt{and} gates to construct a single aggregated condition node $C$. This transformation enables the model to capture the joint condition $C = C_1 \land C_2 \land \ldots \land C_k$ through logical conjunction.

Once the final condition node $C$ is obtained, we apply Equation \ref{eq:conditional_prob} in the same manner as with a single condition node, treating $P(A \mid C)$ as the target. This allows the model to generalize single-condition inference to multi-condition scenarios in a structurally consistent way.

%% file: 04_Exp.tex
\section{Experiments}


\subsection{Dataset Generation}

In most cases, conditioning on a specific node $C$ primarily affects a local subregion of the circuit rather than its entirety. We define this subregion as the \textit{influence area} of $C$, constructed in three steps. First, we identify the fanin cone of $C$. Within this cone, we then locate an ancestor node $P$ that satisfies two conditions: it has multiple fanouts, and it differs from $C$ by no more than 10 logic levels. Finally, starting from $P$, we collect all nodes within its fanout cone, bounded to a depth of 10 logic levels downstream.

We use a dataset of 10,824 AIGs from ITC99~\cite{itc99}, EPFL~\cite{epfl}, and OpenCore~\cite{opencore}, with circuit sizes ranging from 36 to 3,214 gates. To make conditional information explicitly available to the model, we insert \texttt{and} and \texttt{div} gates within each node’s \textit{influence area}. For supervision, we conduct 20,000 random simulations per circuit to record full truth assignments.

\subsection{Experiment Setting}

In the one-round GNN model configuration, both the structural embedding $h^s$ and the functional embedding $h^f$ are set to a dimension of 128. The task head, $\varphi$, has 3 hidden layers with 32 neurons and utilizes the SiLU activation function.

The model is trained for 60 epochs to ensure convergence with a batch size of 512 using a single NVIDIA RTX 4090 GPU. The Adam optimizer~\cite{adam} is used with a learning rate of $10^{-4}$.

\subsection{Performance on Conditional Probability Prediction}

\begin{table}[htbp]
  \centering
  \small
  \begin{tabular}{c|ccc}
    \toprule
    \#Condition Nodes & \#Nodes=928 & \#Nodes=6428 & \#Nodes=17796 \\
    \midrule
    1 & 0.0243 & 0.0372 & 0.0457 \\
    2 & 0.0267 & 0.0401 & 0.0496 \\
    5 & 0.0314 & 0.0479 & 0.0532 \\
    \bottomrule
  \end{tabular}
  \caption{Conditional probability prediction performance (MAE) under varying condition counts and circuit sizes.}
  \label{tab:cond_prob_results}
\end{table}

We train the model on small circuits and evaluate its conditional probability prediction accuracy on both small and large-scale benchmarks. 
In the single-condition setting, the average L1 loss on the validation set reaches as low as \textbf{0.0331}. 
To illustrate generalizability, we select representative circuits of varying sizes. As shown in Table ~\ref{tab:cond_prob_results}, the model maintains low prediction error across all scales, with L1 loss as low as 0.0243 for small circuits and remaining below 0.05 even for circuits with over 17K nodes. 
As the number of conditions increases from 1 to 5, prediction error rises slightly yet remains within acceptable bounds, demonstrating strong robustness. Importantly, inference overhead under multi-condition settings is negligible, thanks to the lightweight design of the aggregator functions of and gates.

\subsection{Ablation Study}
\subsubsection{Effectiveness of \texttt{div} Gate}


\begin{table}
\centering
\small
\begin{tabular}{lcccccc}
\toprule
 & \multicolumn{2}{c}{Moderate Condition} & \multicolumn{2}{c}{Polar Condition} \\
\cmidrule(lr){2-3} \cmidrule(lr){4-5}
\# Cond. & \textbf{\texttt{div} gate} & direct division & \textbf{\texttt{div} gate} & direct division  \\
\midrule
1   &  0.0312 & 0.1173 & 0.0395 &  0.7129  \\
2   &  0.0339 & 0.1495 & 0.0423 &  0.7940  \\
5   &  0.0368 & 0.2030 & 0.0486 &  0.8546  \\
\bottomrule
\end{tabular}
\caption{Ablation study on the effectiveness of using \textbf{\texttt{div} gates} for computing conditional probability (MAE). 
    Here, “\#Cond.” indicates the number of condition nodes involved. }
\label{tab:ablation_div}
\end{table}



Table~\ref{tab:ablation_div} compares our \textbf{\texttt{div} gate}-based model with direct division for conditional probability prediction. Across all settings, \textbf{\texttt{div} gates} yield consistently lower MAE, e.g., reducing error from 0.1173 to 0.0312 with one moderate condition. This highlights their advantage in numerical stability and learning robustness, while direct division suffers from unstable gradients and sensitivity to small denominators.

\subsubsection{Effectiveness of Training Strategy}

\begin{table}
    \centering
    \small
    
    {
    \begin{tabular}{cccc}
        \toprule
        \# Condition Nodes & \textbf{Two-stage} & Only Stage-1 & Only Stage-2 \\
        \midrule
        1 & 0.0331 & 0.0422 & 0.0435 \\
        2 & 0.0352 & 0.0583 & 0.0493 \\
        5 & 0.0387 & 0.0738 & 0.0588 \\
        \bottomrule
    \end{tabular}
    }
    \caption{Ablation study on the effectiveness of our \textbf{two-stage training strategy} (MAE). }
    \label{tab:ablation_twostage}
\end{table}

We evaluate the effectiveness of our two-stage training strategy, where stage-1 learns fine-grained behavior from a 100-pattern simulation and stage-2 captures broader statistical distributions using 20,000-pattern data. As shown in Table~\ref{tab:ablation_twostage}, the two-stage model consistently achieves the lowest L1 loss, outperforming both stage-1 and stage-2 models individually. For instance, with one condition node, L1 loss improves from 0.0422 (stage-1) and 0.0435 (stage-2) to 0.0331 (two-stage).

Additionally, the second-stage model accelerates inference, reducing the time by a factor of 200 compared to stage-1, which needs 200 invocations to simulate the 20,000-pattern workload.

%% file: 05_CSAT.tex
\section{Application to Circuit SAT}
\label{CSAT}



\subsection{Experiment Setting}


We evaluate the proposed methods (see Section~\ref{subsec:problem-definition}) in the context of the Logic Equivalence Checking (LEC) task, one of the most critical CSAT problem determining whether two given circuit designs are logically equivalent. All the experiments, including model inference and SAT solving are conducted on an Intel(R) Platinum 8474C.

Since CDCL SAT solvers exhibit fundamentally different behaviors when solving satisfiable and unsatisfiable instances~\cite{oh2015between}, we conduct separate evaluations of our proposed probability-based phase selection and clause filtering techniques. 
For example, solvers favor variable branching heuristics to quickly find satisfying assignments, but rely on advanced clause learning and prune strategies to prove unsatisfiability. 
To assess the phase selection method, we construct 150 hard satisfabile instances as \texttt{LECSAT} set by applying logic synthesis and minor revisions of different circuits from ForgeEDA dataset~\cite{forgeeda2025}. Circuit pairs are then combined using miter construction via the ABC tool~\cite{ABC}. These instances have an average solving time of 17.39 seconds using \texttt{CaDiCaL}, ensuring that they are sufficiently challenging.
Besides, to evaluate the proposed clause filtering technique, we construct another set of unsatisfiable instances \texttt{LECUNSAT} by generating miters from pairs of datapath circuits.  In these cases, the original circuits are logically equivalent by design, ensuring that the resulting miter circuits are unsatisfiable, which cannot be solved within 20 seconds by \texttt{CaDiCal}.




\subsection{Node Probability for Phase Selection}

    \begin{figure}[ht]
        \centering
        \includegraphics[width=\linewidth]{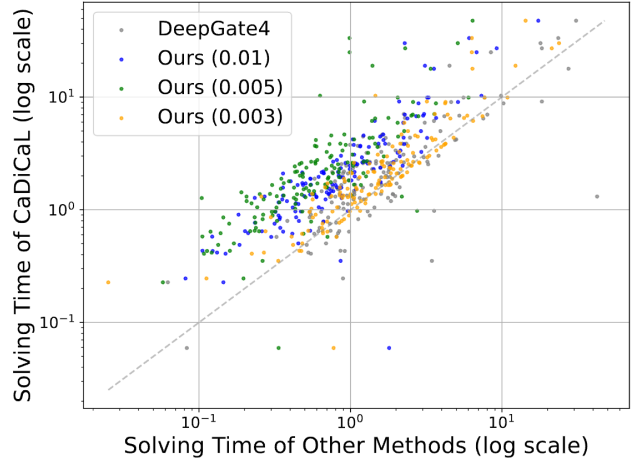}
        \caption{Solving Time on \texttt{LECSAT} (150 SAT cases) with probability-based phase selection(unit: seconds)}
        \label{fig:csat}
    \end{figure}


We use CASCAD to predict condition probability and evaluate our selection method(defined in Section \ref{subsec:problem-definition}) on 150 SAT instances from the \texttt{LECSAT} benchmark under different thresholds \(\tau\). Figure~\ref{fig:csat} compares solving times of our method (x-axis) under various threshold with \texttt{CaDiCaL} (y-axis), both in log scale. We also include the recent circuit representation model DeepGate4~\cite{zheng2025deepgate4} (black dots) as a comparison, which predicts node-level similarity and encourages assigning opposite values to functional similar nodes to trigger conflicts earlier and prune the search.

Most points lie above the diagonal (\(y = x\)), indicating that our method \textbf{outperforms the mainstream phase selection heuristic widely used in modern SAT solvers(has integrated into \texttt{CaDiCaL})}. Compared to DeepGate4, our method consistently achieves lower solving times, demonstrating more effective phase guidance. At \(\tau = 0.005\), our method achieves up to 10\(\times\) speedup, and nearly 5\(\times\) at \(\tau = 0.01\). Points further left imply lower average solving times, highlighting the overall efficiency of our probability-guided strategy across different \(\tau\) values.


In addition, the phase selection method serves as a lightweight classifier for UNSAT instances: since it greatly accelerates SAT solving— if a case remains unsolved within a fixed time budget, it is likely to be UNSAT. This insight allows us to adapt solver parameters dynamically to better handle UNSAT-dominated scenarios(see Appendix~\ref{app:unsat_tuning}), achiving 2× average speedup on \texttt{LECUNSAT} dataset.


\subsection{Clause Probability for Clause Management}

We validate our clause probability metric using a simple protocol. The solver runs until 50,000 conflicts are reached, after which all learnt clauses are extracted. Our \ourmethod model computes each clause’s probability (as defined in Section~\ref{subsec:problem-definition}) and retains only those below a predefined threshold. We apply this filtering in \texttt{CaDiCaL} with varying thresholds \(\tau\), and evaluate performance using the PAR-2 (Par2) score, which penalizes timeouts with twice the cutoff time to reflect both efficiency and robustness.

    \begin{figure}
        \centering
        \includegraphics[width=0.95\linewidth]{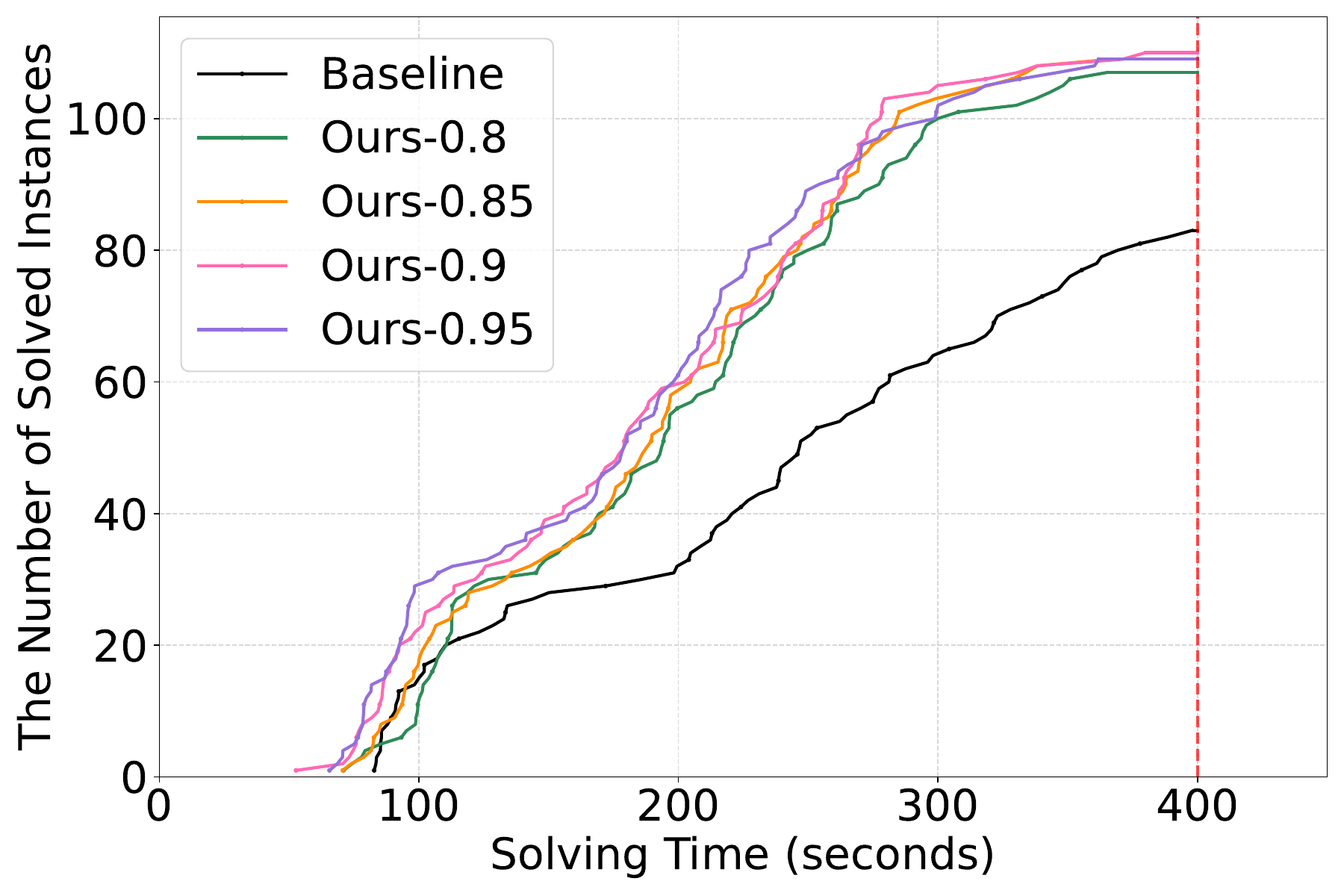}
        \caption{Number of solved instances as a function of solving time using probability-based clause filtering. A total of 130 small case in \texttt{LECUNSAT} dataset. TIMEOUT = 400s.}
        \label{fig:wo_preprocessing}
    \end{figure}

    \begin{table}
    \centering
    \small
    \begin{tabular}{lccc}
    \toprule
    Threshold & Baseline & Ours.Solving & Ours.Overall \\
    \midrule
    0.8  & \multirow{4}{*}{\shortstack[c]{1828.83}} & 1560.72 & 1563.45  \\
    0.85 &                                          & 1406.19 & 1408.03  \\
    0.9  &                                          & 1396.13 & \textbf{1399.34}  \\
    0.95 &                                          & 1429.60 & 1432.48  \\
    \bottomrule
    \end{tabular}
    \caption{Average Par2 of 20 large cases in \texttt{LECUNSAT} using probability-based clause filtering. TIMEOUT = 1000s.}
    \label{tab:large_cases}
    \end{table}




We evaluate our probability-guided clause filtering metric on \texttt{LECUNSAT} dataset (150 instances, split into 130 small and 20 hardest cases). Two configurations are compared: (1) the \texttt{CaDiCaL} baseline with mainstream LBD-based metric, and (2) our method with probability-based filtering. \textbf{Total time} includes both \textbf{solving and model inference time}.

We test thresholds, 0.8, 0.85, 0.9, and 0.95, as shown in Figure~\ref{fig:wo_preprocessing} and Table~\ref{tab:large_cases}.
On 130 small cases, our methods consistently outperform the baseline by solving more instances within the same time limit. 
For the 20 challenging cases, our methods significantly outperform the baseline even when accounting for inference time. 


These results demonstrate that \textbf{our filtering method identifies and preserves more useful clauses than the default LBD heuristic} (see Appendix \ref{app:analysis-clasue}). The optimal threshold of 0.9 reflects a trade-off: low thresholds yield high-quality but sparse clauses, weakening propagation; high thresholds increase quantity but introduce noise. 
\subsection{Effectiveness of Inprocessing Beyond Preprocessing}



\begin{table}
\centering
\small
\begin{tabular}{l c c c}
\toprule
Method & Threshold & Solving & Overall\\
\midrule
Preprocessing Only & - & 13.39 & 13.39 \\
\midrule
\multirow{3}{*}{\makecell[l]{Preprocessing\\+\\ \ourmethod}} 
& 0.01  & 3.26 & 5.01 \\
& 0.005 & 1.04 & \textbf{2.89} \\
& 0.003 & 6.29 & 8.35 \\                                 
\bottomrule
\end{tabular}
\caption{Average Par2 of 150 SAT cases in \texttt{LECSAT} using probability-based phase selection with preprocessing.}
\label{tab:preprocess_inprocess_comparison}
\end{table}

\begin{table}
\centering
\small
\begin{tabular}{l c c c}
\toprule
Setting & Threshold  & Solving & Overall \\
\midrule
Preprocessing Only & -  &  304.42  & 304.42 \\
\midrule
\multirow{4}{*}{\makecell[l]{Preprocessing\\+\\ \ourmethod}}
& 0.8   & 245.61   & 248.82  \\
& 0.85  & 228.98   & 231.76  \\
& 0.9   & 221.50   & \textbf{224.95}  \\
& 0.95  & 223.91   & 226.34  \\
\bottomrule
\end{tabular}
\caption{Average Par2 of 150 UNSAT cases using probability-based clause filtering with preprocessing.}
\label{tab:preprocess_inprocess_clause}
\end{table}

To further validate the distinct effectiveness of our proposed inprocessing framework beyond preprocessing techniques, we compare the two strategies:

\begin{itemize}
    \item \textbf{Preprocessing-only}: We adopt the preprocessing techniques proposed in prior work~\cite{25dac}, 
    to transform the circuit into a solver-friendly format.
    
    \item \textbf{Preprocessing + \ourmethod}: Building on the same preprocessing technique, we integrate it into the inprocessing stage via our CASCAD framework.
\end{itemize}

As shown in Table~\ref{tab:preprocess_inprocess_comparison} and Table~\ref{tab:preprocess_inprocess_clause}, even when strong preprocessing technique has been applied, integrating our inprocessing framework leads to substantial additional speedups. 
For SAT cases, the best setting reduces the average PAR2 from 13.39 to 2.89 (a $4.6\times$ improvement), while for UNSAT cases, our clause filtering achieves a reduction from 304.42 to 224.95.
This highlights the complementary nature of the two techniques, and demonstrates that inprocessing contributes \textit{dynamic guidance} during solving—providing \textit{essential capabilities} that static preprocessing alone cannot achieve.

%% file: 06_Con.tex
\section{Conclusion}


In this work, we propose CASCAD, a novel framework incorporating the inherent circuit information from CSAT instance into dynamic CDCL reasoning. By using a GNN-based model to estimate conditional probabilities of unassigned variable, CASCAD dynamically guides two key CDCL heuristics in the modern SAT solvers: variable phase selection and clause management. Extensive experiments on industry-standard benchmarks demonstrate that CASCAD achieves significant performance gains, reducing solving time by up to 90\% through probability-guided phase selection and achieving an additional 23.5\% reduction via clause filtering in logic equivalence scenarios.

%% file: 07_Appendix.tex
\appendix
\setcounter{table}{0}
\setcounter{figure}{0}
\setcounter{section}{0}
\setcounter{equation}{0}
\renewcommand{\thetable}{A\arabic{table}}
\renewcommand{\thefigure}{A\arabic{figure}}
\renewcommand{\thesection}{A\arabic{section}}
\renewcommand{\theequation}{A\arabic{equation}}

\section{Model Details}
\label{app:model-detail}

\paragraph{Model Architecture.}
We adopt the \textsc{DeepGate4} framework~\cite{zheng2025deepgate4}, where each gate is represented by structural and functional embeddings and updated via self-attention~\cite{vaswani2017attention}.

For each gate type $g \in \{\texttt{and}, \texttt{not}, \texttt{div}\}$, we implement type-specific aggregation and update functions:

\begin{itemize}
    \item \texttt{not.} The aggregator directly propagates the single input and applies a logical negation.
    
    \item \texttt{and / virtual and.} The aggregation function captures conjunction semantics, where attention weights reflect the relative importance of inputs—for example, giving higher weight to controlling inputs that can determine the output value.
    
    \item \texttt{virtual div.} In addition to a two-input aggregator, we account for the asymmetric roles of the inputs (numerator $A_{joint}$ vs.\ denominator $C$) by incorporating positional encodings to distinguish them during message passing.
\end{itemize}

This unified framework enables the model to reason jointly over logical and probabilistic structures.

\paragraph{Level‑by‑Level Propagation.}
Let $\mathcal{L}(v)$ denote the topological level of gate $v$.
For every level $\ell$ (from primary inputs PI to primary outputs PO) the model performs:
\begin{align}
h^{s}_v &\gets \text{Agg}^{s}_{g}\bigl(\{h^{s}_u \mid u \in P(v),\, \mathcal{L}(u) = \ell-1\}\bigr) \\
h^{f}_v &\gets \text{Agg}^{f}_{g}\bigl(\{h^{f}_u, h^{s}_u \mid u \in P(v),\, \mathcal{L}(u) = \ell-1\}\bigr)
\end{align}
where $P(v)$ denotes the set of predecessor nodes of $v$.


\section{Effectiveness of \texttt{div} Gate}
\label{app:div}

The \texttt{div} gate explicitly models conditional probability in the circuit graph via the following equation:
\begin{equation}
P(A \mid C) = \frac{P(A \land C)}{P(C)}\, .
\end{equation}
While this formulation is convenient, the division operation can amplify errors significantly when $P(C)$ is very small (always smaller than 1, as it represents the logic probability of node $C$). To illustrate this behavior, we analyze two representative cases using \textit{ac97\_ctrl.aig} as a case study.

\paragraph{Case~1: Extremely Small $P(C)$.}

\begin{table*}[ht]
    \centering
    \small

    \begin{tabular}{|c|c|c|c|c|c|c|}
        \toprule
        $i$ & \multicolumn{2}{c|}{$P(C)$} & \multicolumn{2}{c|}{$P(A_i \land C)$} & \multicolumn{2}{c|}{$P(A_i \mid C)$} \\
        \midrule
        & ground truth & prediction & ground truth & prediction & ground truth & prediction \\
        \midrule
        1 & 0.01 & 0.00323 & 0.01 & 0.00017 & 1 & 0.05277834 \\
        2 & 0.01 & 0.00323 & 0 & 0.00014 & 0 & 0.04442138 \\
        3 & 0.01 & 0.00323 & 0.01 & 0.00014 & 1 & 0.05133383 \\
        4 & 0.01 & 0.00323 & 0.01 & 0.00016 & 1 & 0.07029346 \\
        5 & 0.01 & 0.00323 & 0 & 0.00009 & 0 & 0.04350258 \\
        \midrule
        $\cdots$ & $\cdots$ & $\cdots$ & $\cdots$ & $\cdots$ & $\cdots$ & $\cdots$ \\
        \midrule
        Overall Error & \multicolumn{2}{c|}{0.00677} & \multicolumn{2}{c|}{0.00786} & \multicolumn{2}{c|}{0.72396} \\
        \bottomrule
    \end{tabular}
    \caption{Performance of the division operation with extremely small $P(C)$. $P(A_i \mid C)$ is calculated by $ \frac{P(A_i \land C)}{P(C)}$}
    \label{tab:div-small}
\end{table*}

In this scenario, when the conditional node rarely evaluates to~1 ($P(C) = 0.01$), even minor absolute errors in $P(A \land C)$ translate into large relative errors in $P(A \mid C)$. Table~\ref{tab:div-small} presents the ground-truth and predicted probabilities for $P(C)$ and a subset of 20 target nodes $A$. The prediction error for $P(C)$ is moderate ($\hat{P}(C) = 0.00323$), yet the resulting conditional probabilities are significantly distorted due to error amplification.

\paragraph{Case~2: Moderate $P(C)$.}

\begin{table*}[ht]
    \centering
    \small
    
    \begin{tabular}{|c|c|c|c|c|c|c|}
        \toprule
        $i$ & \multicolumn{2}{c|}{$P(C)$} & \multicolumn{2}{c|}{$P(A_i \land C)$} & \multicolumn{2}{c|}{$P(A_i \mid C)$} \\
        \midrule
        & gt & pred & gt & pred & gt & pred \\
        \midrule
        1 & 0.42 & 0.4210119 & 0.24 & 0.24765863 & 0.5714286 & 0.58824617 \\
        2 & 0.42 & 0.4210119 & 0.33 & 0.34514248 & 0.7857143 & 0.8197927 \\
        3 & 0.42 & 0.4210119 & 0.31 & 0.3523217 & 0.7380953 & 0.83684504 \\
        4 & 0.42 & 0.4210119 & 0.36 & 0.3949966 & 0.8571429 & 0.93820775 \\
        5 & 0.42 & 0.4210119 & 0.38 & 0.3954696 & 0.9047619 & 0.6914532 \\
        \midrule
        $\cdots$ & $\cdots$ & $\cdots$ & $\cdots$ & $\cdots$ & $\cdots$ & $\cdots$ \\
        \midrule
        Overall Error & \multicolumn{2}{c|}{0.0010119} & \multicolumn{2}{c|}{0.025311} & \multicolumn{2}{c|}{0.101676}\\
        \bottomrule
    \end{tabular}
    \caption{Performance of the division operation with moderate $P(C)$. $P(A_i \mid C)$ is calculated by $ \frac{P(A_i \land C)}{P(C)}$}
    \label{tab:div-medium}
\end{table*}

When $P(C)$ is in a mid-range (e.g., $P(C) > 0.3$), the division operation becomes relatively numerically stable. However, it still amplifies errors in $P(A \mid C)$ to an extent that is practically unacceptable. Table~\ref{tab:div-medium} compares aggregate losses, showing that the overall absolute error remains greater than $0.1$.

In contrast, our method incorporates the \texttt{div} gate during the dataset preparation stage. The probability labels for the \texttt{DIV} gate are computed as the division of the probabilities of its two input nodes, which are tagged with distinct positional markers. \textit{By learning this behavior as a standard gate type}, the model directly internalizes the division operation, eliminating the need to explicitly compute the quotient of joint and marginal probabilities.

\section{Pattern-Based Dataset}

\label{app:pattern}

To bootstrap training, we hope to generate \emph{pattern‑based} traces that approximate full truth tables.
Given a circuit with $m$ primary inputs (PIs), the complete truth table has $2^{m}$ rows—impractical beyond $m\!\approx\!20$.  
Instead, we conduct simulation with 20,000 patterns and uniformly sample \textbf{100} random patterns per circuit per epoch (see Table ~\ref{app:toy_truth_table}) and record the resulting logic probability values for every node.


\begin{table}[H]
\centering
\small
    \begin{tabular}{|c|c|c|c|c|c||c|c|}
    
        \toprule
        Pattern ID & $\!$PI\textsubscript{1}$\!$ & $\!$PI\textsubscript{2}$\!$ & $\!$PI\textsubscript{3}$\!$ & \dots & $\!$PI\textsubscript{$m$}$\!$ &  $\!$Node $u$\!  &  $\!$Node $v$\! \\
        \midrule
        1 & 0  & 1 & 0 & $\dots$ & 1  & 0 & 0\\
        2 & 1  & 0 & 0 & $\dots$ & 0 & 1 & 0 \\
        3 & 0  & 0 & 1 & $\dots$ & 1 & 0 & 1 \\
        \multicolumn{7}{c}{\vdots} \\
        19997 & 0 & 1 & 1 & $\dots$ & 0 &  1 &  1 \\
        19998 & 0 & 1 & 1 & $\dots$ & 0 &  0 &  1 \\
        19999 & 1 & 0 & 1 & $\dots$ & 1 &  1 &  0 \\
        20000 & 0 & 0 & 0 & $\dots$ & 0 &  0 &  1 \\
        \bottomrule
    \end{tabular}
    \caption{Excerpt of a truth table. Each training epoch picks a \emph{fresh} random subset of 100 rows, ensuring the model learns to interpolate between unseen patterns.}
    \label{app:toy_truth_table}
\end{table}

For each node $v$, we compute
\begin{equation}
    \hat P_{\text{rand}}(v)=\frac{1}{100}\sum_{i=1}^{100}\mathbf 1\bigl(v=1 \text{ in row } i\bigr),
\end{equation}
which serves as a supervisory signal encouraging the GNN to capture fine‑grained functional behavior. Because the 100‑row subset changes every epoch, the network is exposed to \emph{thousands} of distinct local views, yielding better generalization than a single large simulation.

\section{Workload-Based Dataset}

\label{app:workload}

While random patterns provide breadth, they fail to capture realistic activity biases. We therefore introduce \emph{workload-based} simulations, where each primary input (PI) is independently assigned a probability value $\rho$ sampled from the set ${0.1, 0.2, \dots, 0.9}$. 
A toy circuit ($c = a \land b$, Fig.~\ref{fig:toy}) is used to illustrate the necessity.

\begin{figure}[H]
    \centering
    \includegraphics[width=0.4\linewidth]{figure/appendix-toy.pdf}
    \caption{Toy circuit $c= a\land b$}
    \label{fig:toy}
\end{figure}


\begin{table}[H]
  \centering
  \small
  \begin{tabular}{cccc}
    \toprule
       $P(a)$ & $P(b)$ & $P(c\!=\!a\land b)$ \\
    \midrule
     0.500 & 0.500 & 0.249  \\
     0.500 & 0.500 & 0.253  \\
     0.100 & 0.400 & 0.040  \\
     0.300 & 0.600 & 0.182  \\
     0.700 & 0.900 & 0.630  \\
    \bottomrule
  \end{tabular}
    \caption{Node probabilities of the toy circuit (Fig.~\ref{fig:toy}) under different PI workload levels $\rho$.}
  \label{tab:workload-toy}
\end{table}


To illustrate this diversity, we simulate two contrasting scenarios (see Table ~\ref{tab:workload-toy}). In the first case, all primary inputs (PIs) are assigned a fixed workload of $\rho=0.5$, under which each internal node converges to a stable logic probability; a representative slice is shown in the first two rows of Table~\ref{tab:workload-toy}. In the second case, each PI's $\rho$ is independently sampled from the set ${0.1, 0.2, \dots, 0.9}$, resulting in a broader range of logic probabilities across internal nodes. This variation exposes the network to a richer set of behaviors and better reflects diverse operational conditions.

\section{Analysis of Probability-Based Metric}
\label{app:analysis-clasue}

In our probability-based clause filtering method, the solving process is paused once the number of conflicts reaches 50{,}000, at which point all learnt clauses are extracted. For better analysis, these clauses are grouped according to their LBD values (1, 2, or 3) and further partitioned based on their predicted probabilities. Each group is then re-inserted into the solver independently, allowing us to evaluate their respective impacts on the remainder of the solving process.

Table~\ref{tab:clause_summary} presents statistics for the learnt clauses under different LBD levels, further divided by probability thresholds (\(<0.8\) and \(\geq 0.8\)). The table reports the number of clauses in each group, along with the corresponding reductions in solving time and decision count.

\begin{table}
\centering
\small
\begin{tabular}{lcccccc}
\toprule
 & \multicolumn{3}{c}{$P < 0.8$} & \multicolumn{3}{c}{$P \geq 0.8$} \\
\cmidrule(lr){2-4} \cmidrule(lr){5-7}
LBD & Num. & Time.Red. & Dec.Red. & Num. & Time.Red. & Dec.Red. \\
\midrule
1 & 1515 & 47.1\% & 55.2\% & 2468 & 5.3\% & 18.9\% \\
2 & 1819 & 21.2\% & 26.6\% & 7142 & -71.0\% & -0.2\% \\
3 & 300 & 6.7\% & 6.5\% & 2728 & -15.6\% & 4.9\% \\
\bottomrule
\end{tabular}
\caption{Statistics of clauses, solving time reduction, and decision count reduction under different LBD values and predicted probabilities.}
\label{tab:clause_summary}
\end{table}

From Table~\ref{tab:clause_summary}, we draw two key conclusions. First, for clauses with the same LBD, retaining those with lower probability is more beneficial for subsequent solving, even though they are fewer in number. In fact, keeping only high-probability clauses can negatively impact solver performance. Second, we observe that the probability distribution of clauses closely matches the distribution of LBD: among clauses with lower LBD, a larger proportion have low probability. We further find that clause probability is a more informative metric than LBD alone. As a baseline, \texttt{CaDiCaL} simply discards all learnt clauses with LBD greater than 2. However, by filtering $LBD=3$ clauses based on probability and retaining only those with low probability, we can still achieve a positive effect on solving performance.

\section{Applying Probability-Based Phase Selection for UNSAT Case}
\label{app:unsat_tuning}

The proposed phase selection method prioritizes variable assignments that are more likely to satisfy the circuit, effectively accelerating SAT solving, reducing solving time by up to 10$\times$ on SAT instances.

Leveraging this property, we regard the method as a lightweight classifier: for a given instance for solving, if the solver fails to find a solution within a predefined time budget under our phase selection heuristic, the instance is likely to be UNSAT. In such cases, we dynamically switch the solver to a configuration tuned specifically for UNSAT problems.

\begin{figure}[H]
    \centering
    \includegraphics[width=0.95\linewidth]{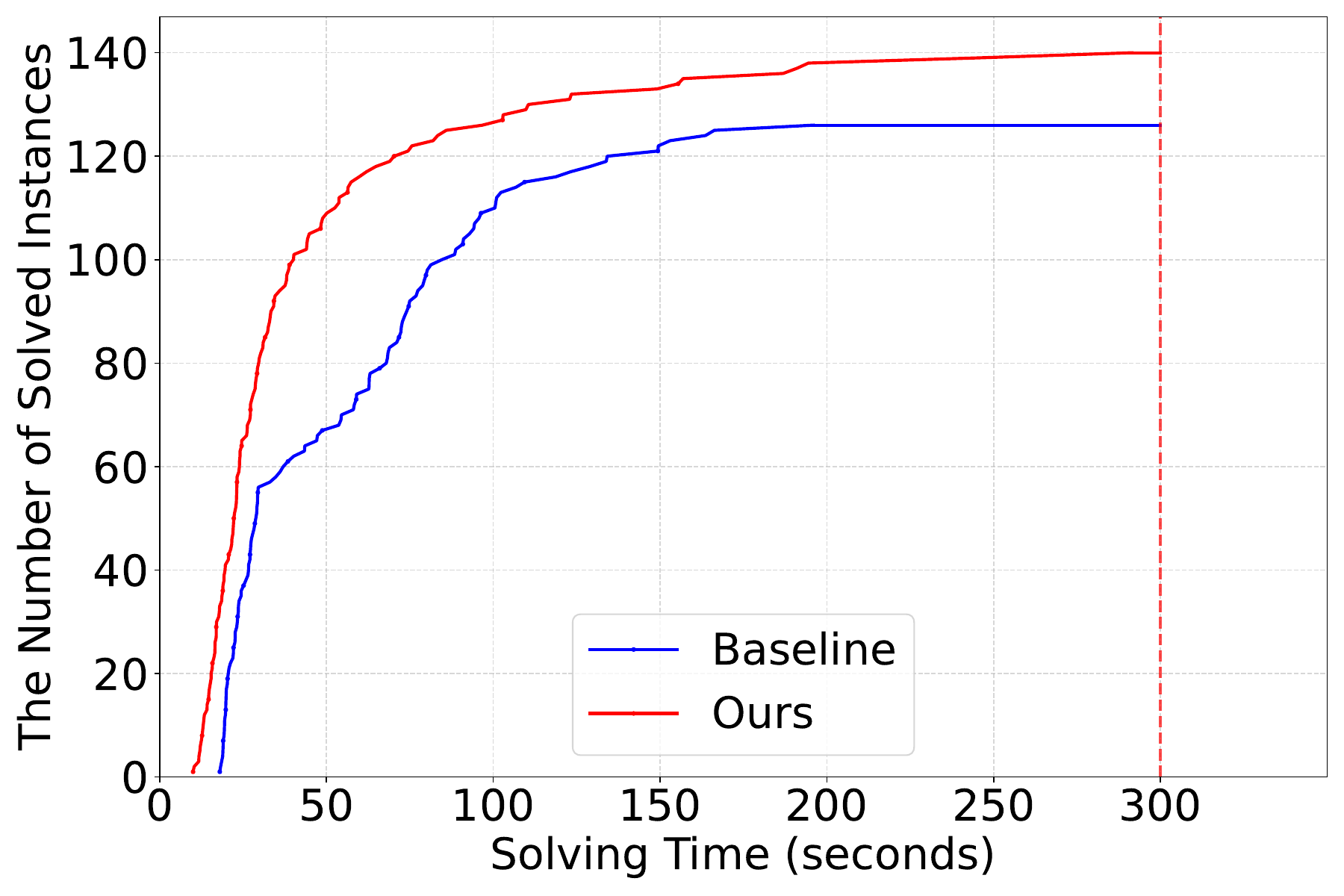}
    \caption{Number of solved instances as a function of solving time using probability-based phase selection. A total of 150 UNSAT case in \texttt{LECSAT} dataset. TIMEOUT = 300s.}
    \label{fig:app_unsat}
    \end{figure}

We implement this strategy on top of \texttt{CaDiCaL}, using a simple timeout-based switching mechanism. We set the time budget as 5 seconds. 
Figure~\ref{fig:app_unsat} summarizes the performance on the \texttt{LECUNSAT} dataset. On this dataset, the average solving time is reduced from 166.26 seconds (raw \texttt{CaDiCaL}) to 80.36 seconds using our adaptive strategy.
As shown in Figure~\ref{fig:app_unsat}, the adaptive strategy consistently reduces solving time across all selected UNSAT-prone cases. 
This demonstrates that early-stage solver behavior under the SAT-oriented phase heuristic provides useful signals for unsatisfiability prediction, enabling effective and low-overhead adaptation to more suitable solving strategies.


